\documentclass[a4,a4wide]{article}
\usepackage{aaai}
\usepackage[utf8]{inputenc}
\usepackage[T1]{fontenc}
\usepackage{amsthm}
\usepackage{amsopn}
\usepackage{amssymb}
\usepackage{amsmath}
\usepackage{tikz}
\usepackage{hyperref}

\pdfpagewidth=\paperwidth
\pdfpageheight=\paperheight
\newcommand*\opp{\textsc{Opp}}
\newcommand*\pro{\textsc{Pro}}
\newcommand*\pl{pl}
\newcommand*\move{\mu}

\newcommand{\context}{C}
\newcommand{\vocabulary}{V}
\newcommand{\foreign}{F}
\newcommand{\cprogram}{P}
\newcommand{\mcs}{\mathcal{C}}

 \newcommand{\arga}{A}

 \newcommand{\argy}{Y}
 \newcommand{\argz}{Z}

\newcommand*\compl[1]{\overline{#1}}

\newcommand{\knowledgebase}{\mathcal{K}}

\newcommand*\query{\mathcal{Q}}

\newcommand*\undefined{-}

\newcommand*\statin{\textsc{In}}
\newcommand*\statout{\textsc{Out}}
\newcommand*\statundec{\textsc{Undec}}
\newcommand*\status[1][s]{\mathit{#1}}

\newcommand*\framework[1][F]{\mathcal{#1}}
\newcommand*\argument[1][A]{\mathit{#1}}
\newcommand*\arguments{\mathcal{A}}
\newcommand{\defeat}{Def}

\newcommand*\conflict[1][C]{\mathit{#1}}
\newcommand*\conflicts[1][C]{\mathcal{#1}}
\newcommand*\vulnerability[1][V]{\mathit{#1}}
\newcommand*\vulnerabilities[1][V]{\mathcal{#1}}
\newcommand*\resolution[1][R]{\mathit{#1}}
\newcommand*\resolutions[1][R]{\mathcal{#1}}
\newcommand*\attack[1][R]{\mathcal{#1}}
\newcommand*\strategy{\sigma}

\newcommand*\argset[1][S]{\mathit{#1}}
\newcommand*\extension[1][E]{\mathcal{#1}}

\newcommand*\eneg{\mathop{\neg}}
\newcommand*\dneg{\mathop{\sim}}
\newcommand*\nimplies{\rightsquigarrow}
\newcommand*\simplies{\rightarrow}
\newcommand*\dimplies{\Rightarrow}
\newcommand*\atom[1][A]{\mathit{#1}}
\newcommand*\literal[1][L]{\mathit{#1}}
\newcommand*\literals[1][L]{\mathcal{#1}}
\newcommand*\infrule[1][r]{\mathit{#1}}
\newcommand*\program[1][P]{\mathcal{#1}}
\newcommand*\strict{\Pi}
\newcommand*\defeasible{\Delta}

\newcommand*\defaults{\literals[D]}
 \newcommand*\lang{\literals[L]}
 
\newcommand{\res}{\mathit{res}}
\newcommand{\con}{\mathit{con}}
\newcommand{\asm}{\mathit{vuls}}

\newcommand{\sem}{\mathit{sem}}

\newcommand*\conc{\textsc{Conc}}
\newcommand*\vuls{\textsc{Vuls}}
\newcommand*\out{\textsc{Output}}

\newcommand{\game}{T}

\newcommand{\entailsk}{\models_{sk}}
\newcommand{\entailcr}{\models_{cr}}

\newtheoremstyle{break}
  {}
  {}
  {\itshape}
  {}
  {\bfseries}
  {.}
  {\newline}
  {}

\theoremstyle{definition}
\newtheorem{definition}{Definition}
\theoremstyle{plain}
\newtheorem{proposition}{Proposition}
\newtheorem{lemma}{Lemma}

\theoremstyle{break}

\theoremstyle{remark}
\newtheorem{example}{Example}

\title{Credulous and Skeptical Argument Games for Complete Semantics in 
  Conflict Resolution based Argumentation
\thanks{This work is supported from the VEGA 
 project no.\,1/1333/12.}}
\author{Jozef Frt\'us\\
Department of Applied Informatics\\
Faculty of Mathematics, Physics, and Informatics\\
Comenius University in Bratislava, Slovakia}

\edef\0{\string\0}
\edef\1{\string\1}
\edef\2{\string\2}
\edef\3{\string\3}

\begin{document}
\nocopyright

\maketitle

\begin{abstract}
 Argumentation is one of the most popular 
 approaches of defining a~non-monotonic 
 formalism and several argumentation based
 semantics were proposed for defeasible
 logic programs.
 Recently, a~new approach based on notions
 of conflict resolutions was proposed,
 however with declarative semantics only.
 This paper gives a~more procedural counterpart 
 by developing skeptical and credulous 
 argument games for complete semantics
 and
 soundness and completeness theorems for
 both games are provided. After that,
 distribution of defeasible logic program
 into several contexts is investigated and
 both argument games are adapted for multi-context system.
\end{abstract}

\section{Introduction}
 Argumentation is successfully applied as an 
 approach of defining non-monotonic formalisms.
 The main advantage of semantics based on 
 formal models of argumentation is its 
 closeness to real humans discussions. 
 Therefore, the semantics can be explained
 also for people not trained in formal logic
 or mathematics.
 
 To capture the knowledge, a~logical language 
 is needed. Usually the language of Defeasible
 Logic Programming (DeLP) is considered, where
 two kinds for rules are distinguished.
 Strict rules represent deductive reasoning: 
 whenever their preconditions hold, we accept the
 conclusion. On the other hand, defeasible rules
 formalize tentative knowledge that can be defeated.
 Several semantics based on argumentation were 
 proposed for defeasible logic programs
 \cite{Prakken1997}, \cite{Garcia2004},
 \cite{Caminada2007}, \cite{Prakken2010},
 \cite{Modgil2011}, \cite{Balaz2013}.
 However, as Caminada and Amgoud \cite{Caminada2007}
 pointed out, careless design of semantics
 may lead to very unintuitive results, such
 as inconsistency of the system (justification
 for both an atom $\atom$ and its negation
 $\eneg \atom$ is provided) or unsatisfying
 of strict rules (system justifies all preconditions,
 but not the conclusion of a~strict rule).
 
 In this paper we take the approach by Bal{\'a}{\v{z}}
 et al. \cite{Balaz2013} as the starting point,
 since it both respects intuitions of logic programming
 and satisfies desired semantical properties.
 In \cite{Balaz2013} notion of conflict resolutions 
 and new methodology of justification of arguments is introduced,
 however only in a~declarative way.
 Our main goal, in this paper, is to give a~more
 procedural counterpart. 
 This is especially useful when dealing
 with algorithms and implementations.
 We adapt skeptical and credulous argument games 
 for complete semantics and prove
 soundness and completeness for both of them,
 what is the main contribution
 of this paper.
 Then we are investigating with distribution of 
 defeasible logic program
 into several contexts (programs) and
 both argument games are adapted for 
 distributed computing.
 This can be useful in ambient intelligence environments,
 where distributed and contextual defeasible reasoning
 is heavily applied.

 The paper is structured as follows:
 first preliminaries of Dung's abstract
 argumentation frameworks and defeasible logic
 programming are introduced. 
 Then the declarative conflict resolution 
 based semantics introduced in \cite{Balaz2013} 
 is recapitulated.
 Argument games are developed and their
 properties are proved in the next section.
 The last section is devoted to contextualization
 of defeasible logic programs.
 
\section{Preliminaries}

\subsection{Argumentation Framework}

\begin{definition}[Abstract Argumentation Framework \cite{Dung1995}]
  An \emph{abstract argumentation framework} is a~pair $\framework = (\arguments, \attack)$ where
  \begin{enumerate}
    \item $\arguments$ is a~set of \emph{arguments}, and
    \item $\attack \subseteq \arguments \times \arguments$ is an \emph{attack}
    relation on $\arguments$.
  \end{enumerate}
\end{definition}

An argument $\argument[A]$ \emph{attacks} an argument $\argument[B]$ if
$(\argument[A], \argument[B]) \in \attack$.
A~set of arguments $\argset$ \emph{attacks} an argument $\argument$ if
an argument in $\argset$ attacks $\argument$.
A~set of arguments $\argset$ is \emph{attack-free}\footnote{Note that we will use the original term ``conflict-free'' in slightly different context.} if $\argset$ does not attack an argument in $\argset$.
A~set of arguments $\argset$ \emph{defends} an argument $\argument$ if
each argument attacking $\argument$ is attacked by $\argset$.
An attack-free set of arguments $\argset$ is \emph{admissible} iff
$\argset$ defends each argument in $\argset$.
 The \emph{characteristic function}
$F_{AF}$ of an argumentation framework $AF = (\arguments, \defeat)$ is a
mapping
$F_{AF}\colon 2^\arguments \mapsto 2^\arguments$ where for all $S \subseteq
\arguments$, $F_{AF}(S)$ is defined as $\{a \in \arguments \mid S \mbox{ defends } a\}$.

\begin{definition}[Extension \cite{Dung1995}]
  An admissible set of arguments $\argset$ is
  \begin{enumerate}
    \item a~\emph{complete extension} iff $\argset$ contains each argument defended by $\argset$.
    \item the \emph{grounded extension} iff $\argset$ is the least complete extension.
    \item a~\emph{preferred extension} iff $\argset$ is a~maximal complete extension.
    \item a~\emph{stable extension} iff $\argset$ attacks each argument which does not belong to $\argset$.
  \end{enumerate}
\end{definition}

We will prove following lemma\footnote{Note that all proofs are presented in the extended version of the paper available at \url{http://dai.fmph.uniba.sk/~frtus/nmr2014.pdf}}, which will be used in procedural formalization
of the grounded semantics. Its intuitive meaning is that an argument $x$
to be in the grounded extension, it can not be defended only by itself.
\begin{lemma}\label{pom}
 Given an argumentation framework $(\arguments, \defeat)$ and
 a~finite ordinal $i$, argument $\arga \in F^{i+1}$ iff 
 for each argument $\argy$ defeating $\arga$, there is an argument
 $\argz \in F^i$ such that $(\argz, \argy) \in \defeat$ and 
 $\argz \neq \arga$.
\end{lemma}

\subsection{Defeasible Logic Program}

An \emph{atom} is a~propositional variable.
A~\emph{classical literal} is either an atom or an atom preceded by classical negation $\eneg$.
A~\emph{default literal} is a~classical literal preceded by default
negation~$\dneg$.
A~\emph{literal} is either a~classical or a~default literal.
By definition $\eneg \eneg \atom$ equals to $\atom$ and $\dneg \dneg \literal$
equals to $\literal$, for an atom $\atom$ and a classical literal
$\literal$.
By $\defaults$ we will denote the set of all default literals.
By convention $\dneg S$ equals to 
$\{\dneg \literal \mid \literal \in S\}$
for any set of literals $S$.

A~\emph{strict rule} is an expression of the form $L_1, \dots, L_n \simplies L_0$ 
where $0 \leq n$, each $L_i$, $1 \leq i \leq n$, is a~literal, and $L_0$ is a~classical literal.
A~\emph{defeasible rule} is an expression of the form $L_1, \dots, L_n
\dimplies L_0$ where $0 \leq n$, each $L_i$, $1 \leq i \leq n$, is
a~literal, and $L_0$ is a~classical literal. 
A~\emph{defeasible logic program} $\program$ 
is a~finite set of of strict rules $\strict$ 
and defeasible rules $\defeasible$.
In the following text we use the symbol $\nimplies$ to denote either strict
or defeasible rule.

\section{Conflict Resolution based Semantics}
Existing argumentation formalisms \cite{Prakken2010,Garcia2004,Prakken1997} are usually defined through five steps.
At the beginning, some underlying logical \emph{language} is chosen for describing knowledge.
The notion of an \emph{argument} is then defined within this language.
Then \emph{conflicts} between arguments are identified.
The resolution of conflicts is captured by an \emph{attack} relation among conflicting arguments.
The \emph{status} of an argument is then determined by the attack relation.

The conflict resolution based approach \cite{Balaz2013} diverge from this methodology.
Instead of attacking a~conflicting argument,
one of the weaker building blocks (called vulnerabilities)
used to construct the argument is attacked.
Specifically, the resolution of a~conflict is either a~default assumption or a~defeasible rule.
The status of an argument does not depend on attack relation between arguments
but on attack relation between conflict resolutions.

Conflict resolution based semantics for the DeLP consists of five steps:
\begin{enumerate}
 \item Construction of arguments on top of the language of defeasible logic programs.
 \item Identification of conflicts between arguments.
 \item Proposing a conflict resolution strategy.
 \item Instantiation of Dung's AFs with conflict resolutions.
 \item Determination of the status of default assumptions, defeasible rules, and arguments
   with respect to successful conflict resolutions.
\end{enumerate}

A vulnerability is a~part of an argument that may be defeated to resolve
a~conflict. It is either a~defeasible rule or a~default literal.
\begin{definition}[Vulnerability]
  Let $\program$ be a~defeasible logic program.
  A~\emph{vulnerability} is a~defeasible rule in $\program$ or a~default literal in $\defaults$.
  By $\vulnerabilities_{\program}$ we will denote the set of all vulnerabilities of $\program$.
\end{definition}

Two kinds of arguments are usually be constructed in the language of defeasible logic programs.
Default arguments correspond to default literals. 
Deductive arguments are constructed by chaining of rules.
The following is a~slightly more general definition,
where a~knowledge base $\knowledgebase$ denotes
literals for which no further backing is needed.

\begin{definition}[Argument]
  \label{def:arg}
  Let $\program = (\strict, \defeasible)$ be a~defeasible logic program.
  An argument $\argument$ for a~literal $\literal$ over a knowledge base 
  $\knowledgebase$ is
  \begin{enumerate}
    \item $[L]$, where $L \in \knowledgebase$
      \begin{align*}
 	\conc(\argument) &= L \\
	\vuls(\argument) &= \{L\} \cap \defaults
      \end{align*}
    \item $[\argument_1, \dots, \argument_n \nimplies \literal]$ where
      each $\argument_i$, $0 \leq i \leq n$, is an argument for a~literal $\literal_i$,
      $\infrule \colon \literal_1, \dots, \literal_n \nimplies \literal$ is a~rule in $\program$.
      \begin{align*}
	    \conc(\argument) &= L\\
	    \vuls(\argument) &= \vuls(\argument_1) \cup \dots \cup \vuls(\argument_n) \cup 
	    (\{\infrule\} \cap \defeasible)
      \end{align*}
  \end{enumerate}
  By $\arguments_{\program}$ we will denote the set of all arguments of $\program$.
\end{definition}
  The typical example of knowledge base within the language
  of defeasible logic programming is the set of
  default literals $\defaults$ and we will not specify
  $\knowledgebase$ until the section about contextual DeLP.
  Therefore, whenever the $\knowledgebase$ is left unspecified, 
  it is implicitly set to $\defaults$.
  Arguments created by chaining of rules will be called \emph{deductive}.
  
\begin{example}
\label{ex:undcons}
Consider the following defeasible logic program $\program$:
\begin{displaymath}
  \begin{array}{rclcrcl}
	&\dimplies& a&\quad&\dimplies& c \\
	&\dimplies& b&\quad&\dimplies& d \\
    a,b&\simplies& h&\quad&c,d&\simplies& \eneg h
  \end{array}
\end{displaymath}
Six deductive arguments can be constructed from $\program$
  \begin{displaymath}
  \begin{array}{rclcrcl}
      \argument_1 &=& [{} \dimplies a]&\quad&
      \argument_4 &=& [{} \dimplies c]\\
      \argument_2 &=& [{} \dimplies b]&\quad&
      \argument_5 &=& [{} \dimplies d]\\
      \argument_3 &=& [\argument_1, \argument_2 \simplies h]&\quad&
      \argument_6 &=& [\argument_3, \argument_4 \simplies \neg h]\\	
  \end{array}
\end{displaymath}
Vulnerabilities of arguments $\argument_3$ are $\argument_6$ are 
$\vuls(\argument_3) = \{\dimplies a, \dimplies b\}$ and
$\vuls(\argument_6) = \{\dimplies c, \dimplies d\}$.
\end{example}

Two kinds of conflicts among arguments may arise, each corresponds to one type of
negation.

\begin{definition}[Conflict]
  Let $\program$ be a~defeasible logic program.
  Arguments $\argument[A], \argument[B] \in \arguments_{\program}$ are \emph{conflicting} iff
  $\argument[A]$ rebuts or undercuts $\argument[B]$ where
  \begin{enumerate}
    \item $\argument[A]$ \emph{rebuts} $\argument[B]$ iff $\argument[A]$ and $\argument[B]$ are deductive arguments and
      $\conc(\argument[A]) = \eneg \conc(\argument[B])$,
    \item $\argument[A]$ \emph{undercuts} $\argument[B]$ iff $\argument[A]$ is a~deductive argument, $\argument[B]$ is a~default argument, and
      $\conc(\argument[A]) = \dneg \conc(\argument[B])$.
  \end{enumerate}
  The set $\conflict = \{\argument[A], \argument[B]\}$ is called a~\emph{conflict}.
  The first kind is called a~\emph{rebutting} conflict and the second kind is called an \emph{undercutting} conflict.
  By $\conflicts_{\program}$ we will denote the set of all conflicts of $\program$.
\end{definition}

Conflicts are resolved by defeating one of the building blocks of conflicting arguments.
Each default assumption or defeasible rule used to construct a conflicting argument
is a possible resolution. Strict rules can not be used as a resolution of any conflict
because they have to be always satisfied.

\begin{definition}[Conflict Resolution]
  Let $\program$ be a~defeasible logic program.
  A~vulnerability $\vulnerability \in \vulnerabilities_{\program}$ is a~\emph{resolution of} a~conflict $\conflict \in \conflicts_{\program}$ if
  $\vulnerability \in \vuls(\conflict)$.
  The pair $\resolution = (\conflict, \vulnerability)$ is called a~\emph{conflict resolution}.
  By $\resolutions_{\program}$ we will denote the set of all conflict resolutions of $\program$.
\end{definition}

In general, each conflict may have more resolutions.
Some of them may be more preferred than others.
The choice of preferred conflict resolutions is always domain dependent.
Some vulnerabilities can be defeated in one domain, but they may as
well stay undefeated in another.
Therefore we allow the user to choose any conflict resolution strategy she might prefer.
\begin{definition}[Conflict Resolution Strategy]
  Let $\program$ be a~defeasible logic program.
  A~\emph{conflict resolution strategy} is a~finite subset $\strategy$ of $\resolutions_{\program}$.
  We say that a~vulnerability $\vulnerability \in \vulnerabilities_{\program}$ is a~$\strategy$-resolution of a~conflict $\conflict \in \conflicts_{\program}$ if
  $(\conflict, \vulnerability) \in \strategy$.
  A~conflict resolution strategy $\strategy$ is \emph{total} iff
  for each conflict $\conflict \in \conflicts_{\program}$ there exists a~$\strategy$-resolution of $\conflict$.
\end{definition}

In existing approaches various conflict resolution strategies
are applied. Examples of default, last-link and weakest-link
conflict resolution strategies are presented in \cite{Balaz2013}.

\begin{example}[Continuation of Example \ref{ex:undcons}]
  \label{ex:undcons2}
  The only conflict in the defeasible logic program $\program$
  is the $\conflict = \{\argument_3, \argument_6\}$.
  Consider following six conflict resolutions.
    \begin{displaymath}
    \begin{array}{rclcrcl}
	\resolution_1 &=& (\conflict, \dimplies a)&\quad&
	\resolution_3 &=& (\conflict, \dimplies c)\\
	\resolution_2 &=& (\conflict, \dimplies b)&\quad&
	\resolution_4 &=& (\conflict, \dimplies d)\\
    \end{array}
  \end{displaymath}
  Then $\strategy = \{\resolution_1\}$, 
  $\strategy' = \{\resolution_i \mid 1 \leq i \leq 4\}$, 
  $\strategy'' = \emptyset$ are examples of
  conflict resolution strategies for $\program$.
  We can see that strategies $\strategy$, $\strategy'$
  are total.
\end{example}

To determine in which way conflicts will be resolved,
Dung's AF is instantiated with conflict resolutions.
The intuitive meaning of a~conflict resolution $(\conflict, \vulnerability)$ is 
``the conflict $\conflict$ will be resolved by defeating the vulnerability
$\vulnerability$''.
The conflict resolution based semantics is built on three levels of attacks: attacks on the
vulnerabilities, attacks on the arguments, and attacks on the conflict
resolutions.
Such an approach is necessary: if a~vulnerability is defeated, so should be all arguments built on
it, and consequently all conflict resolutions respective to the
argument.

\begin{definition}[Attack]
  \label{def:attack}
  A~conflict resolution $\resolution = (\conflict, \vulnerability)$ \emph{attacks}
  \begin{itemize}
    \item a~vulnerability $\vulnerability'$ iff $\vulnerability' = \vulnerability$.
    \item an argument $\argument$ iff $\resolution$ attacks a~vulnerability in $\vuls(\argument)$.
    \item a~conflict resolution $\resolution' = (\conflict', \vulnerability')$ iff either
      \begin{enumerate}
	\item $\vulnerability \neq \vulnerability'$ and $\resolution$ attacks an argument in $\conflict'$ or
	\item $\vulnerability = \vulnerability'$ and $\resolution$ attacks all arguments in $\conflict'$.
      \end{enumerate}
  \end{itemize}
  A~set of conflict resolutions $\argset \subseteq \resolutions_{\program}$
  attacks a~vulnerability $\vulnerability \in \vulnerabilities_{\program}$
  (resp. an argument $\argument \in \arguments_{\program}$ or
  a~conflict resolution $\resolution \in \resolutions_{\program}$) iff
  a~conflict resolution in $\argset$ attacks $\vulnerability$ (resp. $\argument$ or $\resolution$).
\end{definition}

Intuitively, it should not happen that both a~conflict resolution
$\resolution = (\conflict, \vulnerability)$ and
a~vulnerability $\vulnerability$ are accepted.
Therefore, if $\resolution$ is accepted, $\vulnerability$ and all arguments
constructed on top of it should be defeated.
The notion of attack between conflict resolutions formalizes the ideas that
there may be more alternatives how to resolve a~conflict and
a~conflict resolution may resolve other conflicts as well, thus causing other conflict
resolutions to be irrelevant.
The distinction between two kinds of attacks between conflict resolutions 
is necessary to achieve the intended semantics when
dealing with self-conflicting arguments. 
The interested reader is kindly referred to \cite{Balaz2013} 
for demonstrative examples.

\begin{definition}[Instantiation]
  The instantiation for a~conflict resolution strategy $\strategy$
  is an abstract argumentation framework $\framework = (\arguments, \attack)$ where
  \begin{itemize}
    \item $\arguments = \strategy$
    \item $\attack$ is the attack relation on $\strategy$ from the Definition \ref{def:attack}.
  \end{itemize}
\end{definition}

Now thanks to the instantiation we can use the Dung's semantics in order to compute
which vulnerabilities (resp. arguments, conflict resolutions) are undefeated (status $\statin$),
defeated (status $\statout$), or undecided (status $\statundec$).

\begin{definition}[Defense]
  \label{def:defense}
  Let $\strategy$ be a~conflict resolution strategy for a~defeasible logic program $\program$.
  A~set of conflict resolutions $\argset \subseteq \strategy$ \emph{defends} a~vulnerability $\vulnerability \in \vulnerabilities_{\program}$
  (resp. an argument $\argument \in \arguments_{\program}$ or a~conflict resolution $\resolution \in \sigma$) iff
  each conflict resolution in $\strategy$ attacking $\vulnerability$ (resp. $\argument$ or $\resolution$) is attacked by $\argset$.
\end{definition}

\begin{definition}[Status]
  Let $\strategy$ be a~conflict resolution strategy for a~defeasible logic program $\program$ and
  $\extension$ be a~complete extension of the instantiation for $\strategy$.
  The \emph{status of} a~vulnerability $\vulnerability \in \vulnerabilities_{\program}$
  (resp. an argument $\argument \in \arguments_{\program}$ or a~conflict resolution $\resolution \in \sigma$) with respect to $\extension$ is
  \begin{itemize}
    \item $\statin$ if $\extension$ defends $\vulnerability$ (resp. $\argument$ or $\resolution$),
    \item $\statout$ if $\vulnerability$ (resp. $\argument$ or $\resolution$) is attacked by $\extension$,
    \item $\statundec$ otherwise.
  \end{itemize}
  Let $\status \in \{\statin, \statundec, \statout\}$.
  By $\arguments_{\program}^{\status}(\extension)$
  we denote the set of all arguments with the status $\status$ with respect
  to a~complete extension $\extension$.
\end{definition}

The following definitions define actual semantics of the DeLP program $\program$
and entailment relation between a~program $\program$ and a~literal $\literal$.

\begin{definition}[Output]
  Let $\strategy$ be a~conflict resolution strategy for a~defeasible logic program $\program$ and
  $\extension$ be a~complete extension of the instantiation for $\strategy$.
  The \emph{output} of $\extension$ is a~set of literals
   $\out_{\program}(\extension) = \{\literal \in \lang \mid \arguments_{\program}^\statin(\extension) \mbox{ contains an argument for } \literal\}$.
\end{definition}

Note that we will omit default literals in output to improve the legibility.

\begin{definition}[Entailment]
  Let $\strategy$ be a~conflict resolution strategy for a~defeasible logic program $\program$ and
  $\framework$ be the instantiation for $\strategy$.
  Defeasible logic program $\program$ \emph{skeptically} (resp. \emph{credulously})
  \emph{entails} a~literal $\literal$, $\program \entailsk \literal$
  (resp. $\program \entailcr \literal$) iff 
  for each (resp. at least one) complete extension $\extension$  of $\framework$,
  $\literal \in \out_{\program}(\extension)$.
\end{definition}

\begin{example}[Continuation of Example \ref{ex:undcons2}]  
  \label{ex:undcons3}
  Consider the conflict resolution strategy $\strategy'$ from
  Example \ref{ex:undcons2}.
  The instantiation for $\strategy'$ is on the Figure \ref{fig:undcons3}.
  
\begin{figure}[ht]
   \centering
  \begin{tikzpicture}
    \node (R1) {$\resolution_1$};
    \node (R2) [right of=R1] {$\resolution_2$};    
    \node (R3) [below of=R2] {$\resolution_3$};
    \node (R4) [left of=R3] {$\resolution_4$};

    \draw[<->] (R3) to (R4);
    \draw[<->] (R4) to (R2);
    \draw[<->] (R4) to (R1);    
    \draw[<->] (R2) to (R1);
    \draw[<->] (R3) to (R1);
    \draw[<->] (R2) to (R3);   
\end{tikzpicture}
   \caption{The instantiation for the conflict resolution strategy $\strategy'$.}\label{fig:undcons3}
\end{figure}
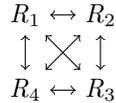
  All conflict resolutions are now exclusive, since to resolve the conflict, 
  it is sufficient to reject only one of the defeasible rules.
  Therefore $\strategy'$ induces the complete graph.
  
  There are five complete extensions $\{\resolution_1\}$, $\{\resolution_2\}$, $\{\resolution_3\}$,
  $\{\resolution_4\}$, $\{\}$ of the instantiation and each of them determine 
  one program output 
  $\{b,c,d,\neg h\}$, $\{a,c,d,\neg h\}$,
  $\{a,b,d,h\}$, $\{a,b,c,h\}$, $\{\}$.  
\end{example}

\section{Procedural Semantics}
 \label{sec:proc}
 In the previous section we recapitulated \cite{Balaz2013} 
 conflict resolution based semantics in the original
declarative way. Although this declarative approach is very elegant and provides nice
algebraic investigations, the more procedural style of semantics is appropriate when
dealing with algorithms and implementations. One can see a parallel in a mathematical logic where
we are similarly interested in a logical calculi (proof theory) which is sound
and complete with respect to defined model-theoretic semantics.
In this section our goal is to define skeptical and credulous
argument games for complete semantics.

  For a~conflict resolution 
  $\resolution = (\{\argument[A], \argument[B]\},\vulnerability)$ we  
  define auxiliary functions which will be frequently used.
    \begin{align*}
    \con(\resolution) &= \{\argument[A], \argument[B]\}\\    
    \res(\resolution) &= \vulnerability\\
    \asm(\resolution) &= (\vuls(A) \setminus \{\vulnerability\}) \cup (\vuls(B) \setminus \{\vulnerability\}) \cup \\
	& \hspace{13pt} (\vuls(A) \cap \vuls(B) \cap \{\vulnerability\})    
  \end{align*}
  $\con(\resolution)$ denotes the conflict and $\res(\resolution)$
  the resolution of a~conflict resolution $\resolution$.
  The meaning of the set of vulnerabilities $\asm(\resolution)$ can
  be explained as following: suppose $\resolution$ is in
  a~conflict resolution strategy $\strategy$ and $\extension$
  is a~complete extension of instantiation for $\strategy$.
  If $\resolution \in \extension$ and all the vulnerabilities in
  $\asm(\resolution)$ have the status $\statin$, then 
  in order to resolve the conflict $\con(\resolution)$, the 
  status of the vulnerability $\res(\resolution)$ is $\statout$.

  Now we characterize the attack between conflict resolutions
  in terms of aforementioned functions.
  This will be useful in proofs for soundness and 
  completeness of argument games.
\begin{proposition}
 Let $\program$ be a~defeasible logic program, $\strategy$ 
 a~conflict resolution strategy and $\resolution = (\conflict, \vulnerability)$,
 $\resolution' = (\conflict', \vulnerability') \in \strategy$ are conflict resolutions.
 Then $\resolution$ attacks $\resolution'$ iff 
 $\res(\resolution) \in \asm(\resolution')$.
\end{proposition}

Argumentation can be seen and thus also
formalized as a discussion of two players. The aim of the first player (called
\emph{proponent}  $\pro$) is to prove an initial argument. The second player is
an \emph{opponent} ($\opp$), what means that her goal is to prevent proponent to
prove the initial argument. Hence a~dispute essentially is a sequence of moves
where each player gives a counterargument to the last stated. 

Proof theory of argumentation is well studied area
and argument games for various semantics were proposed \cite{Modgil2009},
\cite{Prakken1997}. The process of proving a~literal $\literal$
via an argument game, in conflict resolution based setting,
considered in this paper, takes two steps:
\begin{enumerate}
 \item Find an argument $\argument$ with conclusion $\literal$.
 \item Justify all vulnerabilities in $\vuls(\argument)$.
\end{enumerate}

Intuitively, a~move $(\pl, \resolution, \vulnerabilities)$
is a~triple denoting: player $\pl$ claims that
the set of vulnerabilities $\vulnerabilities$ is true 
and resolution $\resolution$ is a~reason for the other player
why her set of vulnerabilities is not justified.

\begin{definition}[Move]
 Let $\strategy$ be a~conflict resolution
 strategy for a~defeasible logic program $\program$.
 A~\emph{move} is a~triple $\move = (\pl, \resolution,
 \vulnerabilities)$, 
 where $\pl \in \{\opp, \pro \}$ denotes the player,
 $\resolution \in \strategy$ is a~resolution and $\vulnerabilities \subseteq 
 \vulnerabilities_{\program}$ is a~set of vulnerabilities. 
\end{definition}
 Now since the very first move in a~dialogue does not 
 counter argue any of the previous move, 
 the resolution $\resolution$ will be left unspecified 
 and in such case we will write
 $(\pl, \undefined, \vulnerabilities)$.
 Convention $\compl{\pro} = \opp$ and
 $\compl{\opp} = \pro$ will be used for denoting the opposite players.
 We say that a~move $(\pl,\resolution,\vulnerabilities)$ attacks 
 a~move $(\compl{\pl},\resolution',\vulnerabilities')$ iff
 $\res(\resolution) \in \vulnerabilities'$.

\begin{definition}[Argument Dialogue]
 A~\emph{dialogue} is a~finite nonempty sequence of
 moves $\move_1,
 \dots,\move_n$, 
 $1 \leq i < n$ where:
 \begin{itemize}
  \item $\pl_i = \pro$ ($\opp$) iff $i$ is odd (even)
  \item $\move_{i+1}$ attacks $\move_{i}$
 \end{itemize} 
\end{definition}

Intuitively, for a given argument, there can be more than one counterargument.
This leads to a tree representation of discussion. Now, since the burden of
proof is on the player $\pro$, proponent proves an initial argument if she wins all
disputes. On the other hand, the burden of attack is on the player $\opp$, meaning
that opponent must ``play'' all possible counterarguments, against $\pro$'s last
argument, forming new branches in a discussion tree.

\begin{definition}[Argument Game]
 Let $\strategy$ be a~conflict resolution
 strategy for a~defeasible logic program $P$.
 An \emph{argument game for an argument} $\argument$ is a~finite tree such that:
 \begin{itemize}
  \item $(\pro, \undefined, \vuls(\argument))$ is the root,
  \item all branches are dialogues,
  \item if move $\move$ played by $\pro$
  is a~node in the tree, then every move 
  $(\opp, \resolution, \asm(\resolution))$ defeating 
  $\move$ is a~child of $\move$.
  \item if $\move$, $\move'$ are any moves played by $\pro$ in $\game$ then $\move$ does not defeat $\move'$.
 \end{itemize} 
\end{definition}

A player wins a~dispute if the counterpart can not 
make any move (give a counterargument).
This can roughly be paraphrased as ``the one who has the last word laughs best''.
Since the burden of proof is on the proponent, $\pro$, 
in order to win, has to win all branches in the game.
On the other hand, for opponent to win an argument game, 
it is sufficient to win at least one branch of the game.
\begin{definition}[Winner]
 A~player $\pl$ \emph{wins a~dialogue} iff
 she plays the last move in it. 
 Player $\pro$ (resp. $\opp$) \emph{wins an argument game} $\game$ iff
 she wins all (resp. at least on of the) branches in the argument game $\game$.
 An argument game is \emph{successful} iff
 it is won by $\pro$. 
\end{definition}

\begin{definition}[Proved Literal]
 Let $\strategy$ be a~conflict resolution
 strategy for a~defeasible logic program $P$.
 A~literal $\literal$ is :
 \begin{itemize}
  \item \emph{proved in an argument
 game $\game$} iff $T$ is a~successful argument game for an argument $\argument$ with $\conc(\argument) = \literal$.
  \item \emph{proved} iff there is
 an argument game $\game$ proving $\literal$. 
 \end{itemize}
\end{definition}

Now we propose two particular argument games and prove 
their soundness and completeness with respect to 
declarative semantics defined in the previous section.

\subsection{Argument Game for Skeptical Complete Semantics}

First we will investigate with skeptical complete semantics
which corresponds to the grounded semantics. 
Since the grounded semantics gives the highest burden of proof
on membership of the extension it defines, the opponent
is allowed to repeat her moves and proponent is not.

\begin{definition}[Skeptical Game]
 An argument game $\game$ is  called \emph{skeptical} iff
 in each branch of $\game$ holds: if $(\pro,\resolution,\vulnerabilities)$,
 $(\pro,\resolution',\vulnerabilities')$ are 
 two moves played by $\pro$, then $\resolution \neq \resolution'$.
\end{definition}

Argument game for skeptical complete semantics is 
sound and complete with respect to
declarative conflict resolution based grounded semantics.

\begin{proposition}
 \label{pro:sk}
 Let $P$ be a~defeasible logic program and
 $\literal$ be a~literal.
 $P \entailsk \literal$ iff 
 $\literal$ is skeptically proved\footnote{
  A~\emph{literal $\literal$ is skeptically proved} iff there is
 an skeptical argument game $\game$ such that $\literal$ is
 proved in $\game$.}.
\end{proposition}
 
 Let demonstrate the skeptical argument game in example.
\begin{example}
 \label{ex:proc1}
 Consider the following defeasible logic program $\program = 
 \{\dimplies a, \dimplies \eneg a\}$ with 
 conflict resolution strategy 
 $\strategy = \{\resolution_1, \resolution_2\}$.
 There are two deductive arguments $\argument_1$,
 $\argument_2$, one conflict $\conflict$ and two
 conflict resolutions $\resolution_1$, $\resolution_2$.
  \begin{displaymath}
  \begin{array}{rclcrcl}
      \argument_1 &=& [{} \dimplies a]&\quad&
      \argument_2 &=& [{} \dimplies \eneg a] \\
      \conflict &=& \{\argument_1, \argument_2\}&\quad&
      & &  \\      
      \resolution_1 &=& (\conflict, \dimplies a)&\quad&
      \resolution_2 &=& (\conflict, \dimplies \eneg a)  \\
  \end{array}
\end{displaymath}
 We would like to skeptically prove literal $a$.
 The skeptical argument game for argument $\argument_1$ is
 on the Figure \ref{fig:proc1}.
 
 \begin{figure}[ht]
   \centering
  \begin{tikzpicture}
    \node (M1) {$\move_1 = (\pro,\undefined,\{\dimplies a\})$};
    \node (M2) [below of=M1] {$\move_2 = (\opp,\resolution_1,\{\dimplies \eneg a\})$};
    \node (M3) [below of=M2] {$\move_3 = (\pro,\resolution_2,\{\dimplies a\})$};    
    \node (M4) [below of=M3] {$\move_4 = (\opp,\resolution_1,\{\dimplies \eneg a\})$};    

    \draw[-] (M2) to (M1);
    \draw[-] (M3) to (M2);
    \draw[-] (M4) to (M3);    
  \end{tikzpicture}
  \caption{The skeptical argument game for argument $\argument_1$.}\label{fig:proc1}
  \end{figure}
  Proponent cannot repeat her move $\move_3$ and therefore
  she loses the game.
\end{example}

\subsection{Argument Game for Credulous Complete Semantics}

Credulous complete semantics corresponds to the preferred semantics,
where an argument can be defended by itself.
Therefore, in credulous game, proponent is allowed to repeat her moves and
opponent is not.

\begin{definition}[Credulous Game]
 An argument game $\game$ is  called \emph{credulous} iff
 in each branch of $\game$ holds: if $(\opp,\resolution,\vulnerabilities)$,
 $(\opp,\resolution',\vulnerabilities')$ are 
 two moves played by $\opp$, then $\resolution \neq \resolution'$.
\end{definition}

Argument game for credulous complete semantics is 
sound and complete with respect to
declarative conflict resolution based preferred semantics.

\begin{proposition}
 \label{pro:cr}
 Let $\program$ be a~defeasible logic program and
 $\literal$ be a~literal.
 $\program \entailcr \literal$ iff 
 $\literal$ is credulously proved\footnote{
  A~\emph{literal $\literal$ is credulously proved} iff there is
 an credulous argument game $\game$ such that $\literal$ is
 proved in $\game$.}.
\end{proposition}

Now we will consider the defeasible logic program $\program$ and
 conflict resolution strategy $\strategy$ from Example
  \ref{ex:proc1} and try to prove literal $a$ credulously. 
\begin{example}[Continuation of Example \ref{ex:proc1}]
 \label{ex:proc2}
 We would like to credulously prove literal $a$.
 The credulous argument game for argument $\argument_1$ is
 on the Figure \ref{fig:proc2}.
 
 \begin{figure}[ht]
   \centering
  \begin{tikzpicture}
    \node (M1) {$\move_1 = (\pro,\undefined,\{\dimplies a\})$};
    \node (M2) [below of=M1] {$\move_2 = (\opp,\resolution_1,\{\dimplies \eneg a\})$};
    \node (M3) [below of=M2] {$\move_3 = (\pro,\resolution_2,\{\dimplies a\})$};    

    \draw[-] (M2) to (M1);
    \draw[-] (M3) to (M2);
  \end{tikzpicture}
  \caption{The credulous argument game for argument $\argument_1$.}\label{fig:proc2}
  \end{figure}
  Opponent cannot repeat her move $\move_2$ and therefore
  the game is successful.
\end{example}

In \cite{Governatori2004,Billington2010} several variants of defeasible logics with procedural semantics are proposed. Repeating an argument for $\pro$ in our approach corresponds to the $\Delta$ proof tag of \cite{Billington2010} and repeating an argument by $\opp$ in our approach corresponds to the $\sigma$ proof tag of \cite{Billington2010}.

\section{Contextual DeLP}
In the previous section we developed a~procedural
semantics based on argument games, now we will 
generalize these ideas to a~distributive setting,
where not only one, but the whole set of
defeasible logic programs is assumed. Each of these
programs may be viewed as a~context (i.e. agent),
which describes the world within its own language 
(i.e. propositional symbols).
Contexts are interconnected into multi-context system
through non-monotonic bridge rules, which import 
knowledge (foreign literals) from other contexts.

Our goal is to adapt the argument games to
multi-context systems and satisfy 
following requirements:
\begin{itemize}
  \item To minimize the necessary communication complexity between
contexts. The conflict between arguments can be decided
in other context, but the structure of arguments should not 
be communicated. 
  \item Contexts provide just distributive computing, they should not
change the semantics. Hence if we look at multi-context system
as a~monolithic program, the output should be the same
as in distributive case.
\end{itemize}

Note that the distributed reasoning is a~very complex task involving also issues of communication protocols and information security.
In this chapter we abstract from this and focus only on the reasoning part.

Distributed computing of semantics is a~hot topic in the
area of multi-agent systems, for Garc{\'i}a and Simari's
\cite{Garcia2004} DeLP a~distributed argumentation framework
was proposed in \cite{Thimm2008}.
Contextual defeasible reasoning is also applied 
in environment of Ambient Intelligence \cite{Bikakis2010},
where devices, software agents and services are supposed
to integrate and cooperate in support of human objectives.

A~vocabulary $\vocabulary$ is a~set of propositional variables.
We say that a~literal is \emph{local}
if its propositional variable is in $\vocabulary$,
otherwise it is \emph{foreign}.
A~\emph{local rule} contains only local literals.
A~\emph{mapping rule} contains local literal in the head and at least one
foreign literal in the body.
A~\emph{contextual defeasible logic program} is a~set of local strict rules,
and local or mapping defeasible rules.

Sometimes we will denote the context 
pertaining to a~foreign literal. 
For example $2 \colon a, c \dimplies b$ means
that foreign literal $a$ is imported from the second context.

\begin{definition}[Context]
A~\emph{context} is a~triple $\context = (\vocabulary, \cprogram, \strategy)$ where
$\vocabulary$ is a~set of propositional variables, $\cprogram$ is a~contextual
defeasible logic program and $\strategy$ is a~conflict
resolution strategy.
\end{definition}

Since, within the one context we do not know
the structure of an argument supporting some
foreign literal, foreign literals cannot be 
used as resolutions of conflicts (their set
of vulnerabilities is empty).

Contextual argument is an argument, where some
of the literals (foreign) do not need a~further backing
and are considered as an import of the knowledge
from the other context.

\begin{definition}[Contextual Argument]
Given a~context $\context = (\vocabulary, \cprogram, \strategy)$ and the set of 
foreign literals $\foreign$,
a~\emph{contextual argument} is an argument over a~knowledge base
$\dneg \vocabulary \cup \foreign$.
The set of all foreign literals contained by an 
argument in a~set of arguments $\arguments$ will be denoted $\foreign(\arguments)$.
\end{definition}

Contextual argument is \emph{foreign} if it is of the form $[L]$,
 where $L$ is a~foreign literal.

Following proposition means that 
foreign literals cannot incorporate a~conflict.

\begin{proposition}
Given a~context $\context = (\vocabulary, \cprogram, \strategy)$ and the set of foreign
literals $\foreign$, a~foreign argument $A$ cannot be in conflict with by any
contextual argument from context $\context$.
\end{proposition}

\begin{definition}[Multi-Context System]
A~\emph{multi-context system}
\footnote{Note that symbol $\context$ was originally used to
denote a~conflict and symbol $\conflicts$ for denoting the set 
of all conflicts. However, the denotation of symbols will always be
clear from the actual text.}
is a~finite nonempty set of contexts $\mcs =
\{\context_1, \dots, \context_n\}$ where $0 < n$,
each $\context_i = (\vocabulary_i, \cprogram_i, \strategy_i)$, $1 \leq i \leq n$, is
a~context
and $\{\vocabulary_1, \dots, \vocabulary_n\}$ is a~partition of the set of all
propositional variables
in $\bigcup_{i=1}^n \cprogram_i$.
\end{definition}
 A~multi context system $\mcs$ is \emph{cyclic} iff there are 
 contexts $\context_1, \context_2, \dots, \context_n$, $n \geq 2$ such that context $\context_i$, $1 \leq i < n$, contains a~mapping rule with a~foreign literal from the context $\context_{i+1}$ and $\context_n$, contains a~mapping rule with a~foreign literal from the context $\context_1$. A~multi context system is \emph{acyclic} iff it is not cyclic.

Sometimes it is useful to look at a~multi-context
system as a~monolithic defeasible logic program and vice
versa.
We say that a~multi-context
system $\mcs = \{\context_1, \dots, \context_n\}$
is a~\emph{contextualization} of 
a~defeasible logic program $\program$ and conflict resolution
strategy $\strategy$ iff $\program = \bigcup_{i=1}^n \cprogram_i$ and
$\strategy = \bigcup_{i=1}^n \strategy_i$. 
The idea of contextualization of a~program or an argument
is illustrated in the following example.
\begin{example}
\label{ex:con}
  Consider the following multi-context system consisting of two 
  contexts
  \begin{center}
    \begin{tabular}{ c | c}
     $\context_1 = (\{a,d,h\}, \cprogram_1, \strategy_1)$
    & $\context_2 = (\{b,c\}, \cprogram_2, \strategy_2)$ \\ \hline
    $\dimplies a$ & $ \dimplies b$ \\
    $\dimplies d$ & $ \dimplies c$ \\    
    $2 \colon b, a \simplies h$ &  \\
    $2 \colon c, d \simplies \eneg h$ &  \\\\        
    $\strategy_1 = \{(\{\argument^{1}_3,\argument^{1}_6\}, \dimplies a)\}$ &
    $\strategy_2 = \emptyset$ \\    
    \end{tabular}
  \end{center}
  
  Six contextual arguments can be constructed in $\cprogram_1$
  \begin{displaymath}
  \begin{array}{rclcrcl}
      \argument^{1}_1 &=& [{} \dimplies a]&\quad&
      \argument^{1}_4 &=& [c]\\
      \argument^{1}_2 &=& [b]&\quad&
      \argument^{1}_5 &=& [{} \dimplies d]\\
      \argument^{1}_3 &=& [\argument^{1}_1, \argument^{1}_2 \simplies h]&\quad&
      \argument^{1}_6 &=& [\argument^{1}_4, \argument^{1}_5 \simplies \neg h]\\	
  \end{array}
\end{displaymath}  
  Two contextual arguments can be constructed in $\cprogram_2$
  \begin{displaymath}
  \begin{array}{rclcrcl}
      \argument^{2}_1 &=& [{} \dimplies b]&\quad&
      \argument^{2}_2 &=& [{} \dimplies c]
  \end{array}
\end{displaymath}  
  
We can see that $\mcs$ is a~contextualization of defeasible
logic program $\program$ and conflict resolution strategy
$\strategy$ from Example \ref{ex:undcons2}.
Similarly, we will define a~notion of
\emph{contextual version of argument} by examples:
arguments $\argument^{1}_1$,
$\argument^{1}_3$, $\argument^{1}_5$, $\argument^{1}_6$ are
(in order) contextual versions of arguments $\argument_1$,
$\argument_3$, $\argument_5$, $\argument_6$, 
but $\argument^{1}_2$, $\argument^{1}_4$ are not contextual 
versions of arguments $\argument_2$, $\argument_4$ in Example \ref{ex:undcons}.
\end{example}

The process of proving a~literal $\literal$ via
an argument game in contextual setting is still 
consisting of two steps:
\begin{enumerate}
 \item Find a~contextual argument $\argument$ with conclusion $\literal$.
 \item Justify all vulnerabilities in $\vuls(\argument)$ and 
 send acceptance queries to contexts pertaining to foreign literals $\foreign(\{\argument\})$.
\end{enumerate}
The second step means that whenever a~player $\pl$
plays in a~dialogue a~move $\move$, not only all vulnerabilities of
$\move$ but also all foreign literals occurring in 
$\move$ must be justified in order to $\pl$ will be the winner.

It is not hard to see that support dependency through foreign literals
may be cyclic in a~multi-context system.
For example context $\context_1$ may use a~foreign literal from context $\context_2$
and vice versa.
Therefore we have to take care of termination of the
queries to other contexts.

\begin{example}
  Consider the following multi-context system consisting of two 
  contexts, each using foreign literal from the other 
  context.
  \begin{center}
    \begin{tabular}{ c | c}
    Context 1 & Context 2 \\ \hline
    $\dimplies a$ & $1 \colon a \dimplies \eneg b$ \\
    $2 \colon b \dimplies \eneg a$ & $\dimplies b$ \\\\
    $\strategy_1 = \{\resolution_1 = (\conflict_1, \dimplies a)\}$ &
    $\strategy_2 = \{\resolution_2 = (\conflict_2, \dimplies b)\}$ \\    
    \end{tabular}
  \end{center}
Where conflict $\conflict_1 = \{[\dimplies a], [2 \colon b \dimplies \eneg a]\}$ and
conflict $\conflict_2 = \{[1 \colon a \dimplies \eneg b], [\dimplies b]\}$.  

Consider now query about credulous acceptance of literal $a$.
There is only one rule deriving $a$ and the only conflict
resolution $\resolution_1$ defeating 
it. Recall the intuitive meaning of conflict resolution in distributive 
setting:
If the vulnerability $\{b \dimplies \eneg a\}$
and foreign literal $b$ are accepted, rule $\dimplies a$
is defeated. Defeasible rule $b \dimplies \eneg a$ is 
not a~resolution of any conflict so its trustworthiness
is not a~subject of dispute. Now the query about acceptance
of the foreign literal $b$ is given to the Context 2.
The process of proving $b$ in Context 2 is similar, therefore
we skip details and only remark that query about acceptance
of the foreign literal $a$ is given back to the Context 1.
We can see that naive adaptation of argument games 
may lead to infinite sending of queries between contexts 
which have cyclic support dependency.
\end{example}

To overcome problem illustrated in the previous example, 
from now on in this paper we investigate with acyclic multi-context systems
only and more general cases are left for the future work.

Now we will define notions for contextual proving and argument games.
Contextual argument game is an argument game $\game$
accompanied with a~query function $\query$
defining queries for every move in $\game$.
Intuitively, a~query is a~foreign literal
that needs to proved in other context.

\begin{definition}[Contextual Argument Game]
 Let $\context$ be a~context and $\move$ be
 a~move $(\pl, \resolution, \vulnerability)$.
 A~\emph{contextual argument game for a~contextual argument} $\argument$ 
 is a~pair $(\game,\query)$, where
 $\game$ is an argument game for $\argument$ and 
 $\query$ is a~query function
 \begin{displaymath}
    \query(\move) = \left\{ \begin{array}{ll}
    \foreign(\{\argument\}) & \textrm{if $\move$ is the root of the tree}\\
    \foreign(\con(\resolution)) & \textrm{otherwise}
    \end{array} \right.
 \end{displaymath} 
 assigning queries for each move.
\end{definition}
We say that a~contextual argument game for a~literal $\literal$
is a~contextual argument game for a~contextual argument $\argument$
with $\conc(\argument) = \literal$. Given a~query 
function $\query$, the set of all foreign literals, played
by a~player $\pl$ in a~contextual argument game
$(\game, \query)$, will be denoted by $\query(\pl)$.

Contextual skeptical and credulous games respect
conditions of move repetitions. That is,
in contextual skeptical (credulous) game, opponent (proponent) is allowed
to repeat her moves and proponent (opponent) is not.
However, since parts of the argument game can be queried
to another contexts, we have to take care that
requirements of (non)repetitions of moves are 
satisfied also there.
Realize that each time a~query about foreign literal $\foreign$
to other context $\context'$ is 
sent from a~move $(\pl,\resolution,\vulnerabilities)$ in
an argument game $\game$,
no matter whether $\pl$ is proponent or opponent,
the argument game for $\foreign$ in context $\context'$
will be started by proponent. Therefore, if $\pl$ 
is $\pro$, the semantics of argument game in context 
 $\context'$ does not change. 
 On the other hand, if $\pl$ is
is $\opp$, the semantics of argument game in context $\context'$
will switch in order
to keep the requirements of (non)repetitions of moves.

This leads into two mutually recursive definitions of
skeptical and credulous contextual argument games.
Note however that the recursion is well-founded (always
terminates) since we are considering multi-context
systems with acyclic support dependency only.

\begin{definition}[Contextual Skeptical Game]
 Let $\move$ be a~move $(\pl, \resolution, \vulnerabilities)$. 
 A~contextual argument game 
 $(\game,\query)$ is called \emph{skeptical} iff
 \begin{itemize}
  \item $\game$ is skeptical game and
  \item for each move in $\game$ with
  $\query(\move) \neq \emptyset$ 
  there is a~$\sem(\move)$ contextual argument game, where   
  
  \begin{displaymath}
      \sem(\move) = \left\{ \begin{array}{ll}
      \textrm{skeptical} & \textrm{if $\pl = \pro$}\\
      \textrm{credulous} & \textrm{otherwise}
      \end{array} \right.
  \end{displaymath} 
 defines the acceptance semantics for queries.  
 \end{itemize}  
\end{definition}

\begin{definition}[Contextual Credulous Game]
 Let $\move$ be a~move $(\pl, \resolution, \vulnerabilities)$. 
 A~contextual argument game 
 $(\game,\query)$ is called \emph{credulous} iff
 \begin{itemize}
  \item $\game$ is credulous game and
  \item for each move in $\game$ with
  $\query(\move) \neq \emptyset$ 
  there is a~$\sem(\move)$ contextual argument game, where   
  
  \begin{displaymath}
      \sem(\move) = \left\{ \begin{array}{ll}
      \textrm{credulous} & \textrm{if $\pl = \pro$}\\
      \textrm{skeptical} & \textrm{otherwise}
      \end{array} \right.
  \end{displaymath} 
 defines the acceptance semantics for queries.  
 \end{itemize} 
\end{definition}

Recall that player $\pl$, in order to be the winner,
has to justify not only all
the vulnerabilities played by her, but also
all $\pl$'s queries have to successful.
Hence, although player does not play the last 
move in a~dialogue, she can still be a~winner
if a~query of the second player is not justified.

Again, the definition is recursive but the
assumption of acyclicity guarantees its 
termination.

\begin{definition}[Contextual Winner]
 Let $(\game, \query)$ be a~contextual
 argument game.
 A~player $\pl$ \emph{wins a~dialogue 
 in contextual argument game} $(\game, \query)$ iff
 \begin{itemize}
  \item all contextual argument games
  for literals in $\query(\pl)$ are successful and
  \item at least one of the following holds:
    \begin{itemize}
      \item $\pl$ plays the last move in the dialogue, or      
      \item at least one of the contextual argument game
      for literals in $\query(\compl{\pl})$ is not successful.
    \end{itemize}   
 \end{itemize}
 A~player $\pro$ (resp. $\opp$) \emph{wins a~contextual argument game} iff
 she wins all (resp. at least one of the) branches in the~contextual argument game.
 A~contextual argument game is \emph{successful} iff
 it is won by $\pro$.
\end{definition}

\begin{definition}[Contextually Proved Literal]
 Let $\mcs$ be a~multi-context system and $\context \in \mcs$ be a~context.
 A~literal $\literal$ is (skeptically, resp. credulously) \emph{proved} in:
 \begin{itemize}
  \item a~contextual argument game $(\game,\query)$ iff
  there is a~contextual argument $\argument$ with
 $\conc(\argument) = \literal$, $\game$ is an (skeptical, resp. credulous) argument game for $\argument$ and $(\game,\query)$ is successful.
  \item a~context $\context$ iff
 $\context = (\vocabulary, \cprogram, \strategy)$, $\literal \in
 \vocabulary$ and there is
 a~contextual argument game (skeptically, resp. credulously)
 proving $\literal$.
  \item a~multi-context system $\mcs$ iff there is
 a~context $\context$ such that 
 $\literal$ is (skeptically, resp. credulously) proved in $\context$. 
 \end{itemize} 
\end{definition}

One of our goals was that contextualization of 
a~program provides just a~distributive computing and
should not change its output. 
The following proposition claims that we are successful
by achieving it.

\begin{proposition}
 Let $\mcs$ be an acyclic contextualization of a~defeasible
 logic program $P$ and $\literal$ be a~literal.
 \begin{enumerate}
  \item $P \entailsk \literal$ iff $\literal$ is
  skeptically proved in $\mcs$.
  \item $P \entailcr \literal$ iff $\literal$ is
  credulously proved in $\mcs$.  
 \end{enumerate}

\end{proposition}

Distribution of argument games is demonstrated in example.

\begin{example}
  \label{ex:con3}
  Consider the following multi-context system consisting of two 
  contexts
  \begin{center}
    \begin{tabular}{ c | c}
     $\context_1 = (\{a\}, \cprogram_1, \strategy_1)$
    & $\context_2 = (\{b\}, \cprogram_2, \strategy_2)$ \\ \hline
    $\dimplies a$ & $ \dimplies b$ \\
    $2 \colon b \dimplies \eneg a$ & $ \dimplies \eneg b$ \\\\        
    $\strategy_1 = \{(\{\argument^{1}_1,\argument^{1}_3\}, \dimplies a)\}$ &
    $\strategy_2 = \{(\{\argument^{2}_1,\argument^{2}_2\}, \dimplies b)\}$ \\    
    \end{tabular}
  \end{center}
  
  Three contextual arguments can be constructed in $\cprogram_1$
  \begin{displaymath}
  \begin{array}{rclcrcl}
      \argument^{1}_1 &=& [{} \dimplies a]&\quad&
      \argument^{1}_2 &=& [b]\\
      \argument^{1}_3 &=& [\argument^{1}_2 \dimplies \eneg a]&\quad&
        & &
  \end{array}
\end{displaymath}  
  Two contextual arguments can be constructed in $\cprogram_2$
  \begin{displaymath}
  \begin{array}{rclcrcl}
      \argument^{2}_1 &=& [{} \dimplies b]&\quad&
      \argument^{2}_2 &=& [{} \dimplies \eneg b]
  \end{array}
\end{displaymath} 

The contextual argument game $\game$ (both skeptical and credulous) is on the Figure 
\ref{fig:con31}, the contextual game $\game'$ for query $b$ is  on the Figure 
\ref{fig:con32}.

 \begin{figure}[ht]
   \begin{center}
  \begin{tikzpicture}
    \node (M11) {$\move^{1}_1 = (\pro,\undefined,\{\dimplies a\})$};
    \node (M12) [below of=M11] {$\move^{1}_2 = (\opp,\resolution_1,\{2 \colon b \dimplies \eneg a\})$, $\query(\move^{1}_2) = \{b\}$};

    \draw[-] (M12) to (M11);
  \end{tikzpicture}
  \caption{The contextual argument game for literal $a$ in context $\context_1$.}\label{fig:con31}
  \end{center}  
  \end{figure}  

  \begin{figure}[ht]
  \begin{center}
  \begin{tikzpicture}
     \node (M21) [right of=M12] {$\move^{2}_1 = (\pro,\undefined,\{\dimplies b\})$};    
     \node (M22) [below of=M21] {$\move^{2}_2 = (\opp,\resolution_2,\{\dimplies \eneg b\})$};    

     \draw[-] (M22) to (M21);    
  \end{tikzpicture}  
    \caption{The contextual argument game for a~query $b$ in context $\context_2$.}\label{fig:con32}
  \end{center}  
  \end{figure}
    
  Although the proponent did not play the last move in 
  $\game$, she is still winner, since the query about 
  foreign literal $b$ was not successful.
\end{example}

\section{Conclusion}
We have developed a~procedural conflict resolution based semantics by adaptation
of skeptical and credulous argument games for complete semantics. 
The soundness and completeness properties for both type of games
are proved, what is the main contribution of this paper.
At the end we have showed how the semantics of 
defeasible logic program can be computed in a~distributive fashion
and both skeptical and credulous argument games 
were modified for multi-context systems.
However, only multi-context systems with acyclic
support dependency have been considered and the more general 
cases were left for the future work.


\bibliographystyle{aaai}
\bibliography{nmr2014}

\begin{thebibliography}{}

\bibitem[\protect\citeauthoryear{Bal{\'a}{\v{z}}, Frt{\'u}s, and
  Homola}{2013}]{Balaz2013}
Bal{\'a}{\v{z}}, M.; Frt{\'u}s, J.; and Homola, M.
\newblock 2013.
\newblock Conflict resolution in structured argumentation.
\newblock In {\em Proceedings of the 19th International Conference on Logic for
  Programming, Artificial Intelligence, and Reasoning}.

\bibitem[\protect\citeauthoryear{Bikakis and Antoniou}{2010}]{Bikakis2010}
Bikakis, A., and Antoniou, G.
\newblock 2010.
\newblock {Defeasible Contextual Reasoning with Arguments in Ambient
  Intelligence}.
\newblock {\em IEEE Transactions on Knowledge and Data Engineering}
  22(11):1492--1506.

\bibitem[\protect\citeauthoryear{Billington \bgroup et al\mbox.\egroup
  }{2010}]{Billington2010}
Billington, D.; Antoniou, G.; Governatori, G.; and Maher, M.
\newblock 2010.
\newblock An inclusion theorem for defeasible logics.
\newblock {\em ACM Trans. Comput. Logic} 12(1):6:1--6:27.

\bibitem[\protect\citeauthoryear{Caminada and Amgoud}{2007}]{Caminada2007}
Caminada, M., and Amgoud, L.
\newblock 2007.
\newblock {On the evaluation of argumentation formalisms}.
\newblock {\em Artificial Intelligence} 171(5-6):286--310.

\bibitem[\protect\citeauthoryear{Dung}{1995}]{Dung1995}
Dung, P.~M.
\newblock 1995.
\newblock {On the acceptability of arguments and its fundamental role in
  nonmonotonic reasoning, logic programming and n-person games}.
\newblock {\em Artificial Intelligence} 77(2):321--357.

\bibitem[\protect\citeauthoryear{Garc\'{\i}a and Simari}{2004}]{Garcia2004}
Garc\'{\i}a, A.~J., and Simari, G.~R.
\newblock 2004.
\newblock {Defeasible logic programming: an argumentative approach}.
\newblock {\em Theory and Practice of Logic Programming} 4(2):95--138.

\bibitem[\protect\citeauthoryear{Governatori \bgroup et al\mbox.\egroup
  }{2004}]{Governatori2004}
Governatori, G.; Maher, M.~J.; Antoniou, G.; and Billington, D.
\newblock 2004.
\newblock Argumentation semantics for defeasible logic.
\newblock {\em J. Log. and Comput.} 14:675--702.

\bibitem[\protect\citeauthoryear{Modgil and Caminada}{2009}]{Modgil2009}
Modgil, S., and Caminada, M.
\newblock 2009.
\newblock Proof theories and algorithms for abstract argumentation frameworks.
\newblock In Rahwan, I., and Simari, G., eds., {\em Argumentation in Artificial
  Intelligence}. Springer Publishing Company Incorporated.
\newblock  105--129.

\bibitem[\protect\citeauthoryear{Modgil and Prakken}{2011}]{Modgil2011}
Modgil, S., and Prakken, H.
\newblock 2011.
\newblock {Revisiting Preferences and Argumentation}.
\newblock In {\em Proceedings of the Twenty-Second International Joint
  Conference on Artificial Intelligence},  1021--1026.
\newblock AAAI Press.

\bibitem[\protect\citeauthoryear{Prakken and Sartor}{1997}]{Prakken1997}
Prakken, H., and Sartor, G.
\newblock 1997.
\newblock {Argument-based extended logic programming with defeasible
  priorities}.
\newblock {\em Journal of Applied Nonclassical Logics} 7(1):25--75.

\bibitem[\protect\citeauthoryear{Prakken}{2010}]{Prakken2010}
Prakken, H.
\newblock 2010.
\newblock {An abstract framework for argumentation with structured arguments}.
\newblock {\em Argument \& Computation} 1(2):93--124.

\bibitem[\protect\citeauthoryear{Thimm and Kern-Isberner}{2008}]{Thimm2008}
Thimm, M., and Kern-Isberner, G.
\newblock 2008.
\newblock A distributed argumentation framework using defeasible logic
  programming.
\newblock In {\em Proceedings of the 2008 Conference on Computational Models of
  Argument: Proceedings of COMMA 2008},  381--392.
\newblock Amsterdam, The Netherlands, The Netherlands: IOS Press.

\end{thebibliography}

\end{document}